\RequirePackage{fix-cm}
\documentclass[preprint,review]{elsarticle}
\usepackage{prletters}

\usepackage{amsmath,amssymb,amsfonts}
\newcommand{\norm}[1]{\left\lVert#1\right\rVert}
\usepackage{hyperref}
\usepackage{cleveref}
\usepackage{mathtools}
\usepackage{algpseudocode}
\usepackage{algorithm}
\usepackage{graphicx}
\usepackage{textcomp}
\usepackage{multirow}
\usepackage{xcolor}
\usepackage{makecell}
\usepackage{multicol}
\usepackage{colortbl}

\usepackage{multirow}
\usepackage{tabularx}
\newcolumntype{L}{>{\raggedright\arraybackslash}X}
\usepackage{threeparttable}

\usepackage{graphicx,placeins}
\usepackage[font=small,skip=4pt]{caption}

\begin{document}

\begin{frontmatter}

\title{Enhancing Pattern Classification in Support Vector Machines through Matrix Formulation}




\cortext[cor1]{Corresponding author}
\author[label1]{Sambhav Jain \corref{cor1}}
\ead{sambhavjain0712@gmail.com}

\author[label1]{Reshma Rastogi (nee Khemchandani)}
\address[label1]{Department of Computer Science, South Asian University, New Delhi-110068}



\begin{abstract}
Support Vector Machines (SVM) have gathered significant acclaim as classifiers due to their successful implementation of Statistical Learning Theory. However, in the context of multiclass and multilabel settings, the reliance on vector-based formulations in existing SVM-based models poses limitations regarding flexibility and ease of incorporating additional terms to handle specific challenges. To overcome these limitations, our research paper focuses on introducing a matrix formulation for SVM that effectively addresses these constraints. By employing the Accelerated Gradient Descent method in the dual, we notably enhance the efficiency of solving the Matrix-SVM problem. Experimental evaluations on multilabel and multiclass datasets demonstrate that Matrix SVM achieves superior time efficacy while delivering similar results to Binary Relevance SVM.\\

Moreover, our matrix formulation unveils crucial insights and advantages that may not be readily apparent in traditional vector-based notations. We emphasize that numerous multilabel models can be viewed as extensions of SVM, with customised modifications to meet specific requirements. The matrix formulation presented in this paper establishes a solid foundation for developing more sophisticated models capable of effectively addressing the distinctive challenges encountered in multilabel learning.

\end{abstract}


\end{frontmatter}


\section{Introduction}
\label{Introduction}

Support Vector Machines is an esteemed binary classifier that has gained recognition due to its effective implementation of Statistical Learning Theory. The combination of practical usability and robust theoretical foundations has made SVMs successful in a wide range of real-world applications, including bioinformatics \cite{yin2015scene}, scene classification \cite{subasi2013classification}, and power applications \cite{hao2010partial}. Furthermore, the influence of SVM has sparked significant interest among researchers, leading to the development of novel non-parallel classifiers such as Generalised Eigenvalue Proximal Support Vector Machines (GEPSVM) \cite{mangasarian2005multisurface} and Twin Support Vector Machines (TSVM) \cite{khemchandani2007twin}. These advancements have opened up new avenues for research and exploration in the field of machine learning. For example, Parametric Non-Parallel Support Vector Machine \cite{jain2022parametric} is an interesting classifier that can model parallel or non-parallel hyperplanes depending upon the choice of the hyperparameter, in addition to showing functional analogy to Pin-SVM.

SVM has been extended to handle the challenges of multiclass and multilabel settings, such as Multiclass Kernel-based Vector Machines \cite{crammer2001algorithmic} and multilabel methods like Multilabel Relationship Learning \cite{zhang2013multilabel}, Rank SVM \cite{elisseeff2001kernel}. Authors in \cite{shajari2020unified} present a Unified SVM approach term as One vs None (OVN) to tackle pattern classification under both settings. Additionally, the Binary Relevance (BR) technique \cite{boutell2004learning} is commonly employed to address classification tasks in both scenarios.

However, these existing models typically rely on vector-based formulations, hindering them from unlocking their full potential. The vector-based formulations have increased training time since each problem is handled independently, for example. Furthermore, the complexity associated with vector-based formulations presents a significant challenge when it comes to incorporating additional functionalities or making modifications to address specific issues. For example, efficiently handling classification in multilabel learning with missing labels, while also effectively leveraging label correlations, becomes a daunting task within vector-based formulations.
In essence, using vector notation-based models present limitations regarding flexibility and ease of modification to incorporate additional terms for handling specific challenges.

In this research, we focus on developing a Matrix formulation for SVM to effectively deal with the aforementioned issues. Comparison between Matrix SVM and BR-SVM in the same environment for multiclass, multilabel problems demonstrates a superior training time advantage of the former with similar results. Both are solved in the dual, using Accelerated Gradient Descend (AGD) for faster convergence. To add more, this paper lays a solid foundation for developing more sophisticated models to effectively address the unique challenges in multi-label learning.

The present work is discussed in the following sections. Notations are discussed in \cref{Notations}. Our proposed model is introduced in \cref{Proposed Model} and subsequently discusses its connection with the existing approaches. The experimental results are reported in \cref{Experiments}, and finally, \cref{Conclusions} concludes the paper.

\section{Notations}
\label{Notations}
Consider a set of $n$ instances, $X=$ $\left\lbrace x_{1}, x_{2}, ...., x_{n} \right\rbrace$ each in a $(d+1)$ dimensional space, i.e., $x_{i}$ $\in$ $\mathbb{R}^{d+1}$, $i = 1, 2,.....,n$ be associated with $m$ labels or classes. Note that the data matrix $X$ is augmented by a column vector of ones for simpler calculations. The corresponding label vector $ y_i$ $\in$ \{1, -1\}$^{m}$, determines association of x$_{i}$ to each of these classes. In general, the data set, $X$ $\in$ $\mathbb{R}^{n \times (d+1)}$, $Y$ is the ground truth label matrix, $Y$ $\in$ \{1, -1\}$^{n \times m}$. 
$y_{i,j}\,=\,1$ means that sample $x_{i}$ is associated to class $c_{j}$, similarly $y_{ij}\,=\,-1$ indicates that sample $x_{i}$ is not associated to class $c_{j}$. In multiclass scenario, a sample can belong to only one class, whereas in multilabel setting, a sample may be associated with multiple classes (better known as labels). $X_{n_{t}}$ $\in$ $\mathcal{R}^{n_{t} \times (d+1)}$ are the testing samples.

\section{Proposed Model}
\label{Proposed Model}

\subsection{Motivation}
\label{Motivation}
There exist numerous techniques that address multiclass, multilabel problems using vector-based optimisation. However, adopting a matrix-based approach to classification offers greater flexibility and facilitates tailored modifications to tackle specific challenges. 
Moreover, in the context of multiclass multilabel classification, employing a matrix formulation to solve a single unified optimization task is more time-efficient compared to solving multiple optimizations using binary relevance.
Introducing a Matrix SVM enables exploring various insights that may remain hidden in a vectorized format, as discussed in the \cref{Connection with existing approaches}. Additionally, when it comes to framing a multilabel optimisation problem, the widely employed least squares loss and squared hinge loss have dominated the field. The non-smooth nature of hinge loss has attracted limited attention in multilabel learning. Nevertheless, we effectively leverage the hinge loss by solving the Matrix SVM in its dual form.

\subsection{Formulation}
The matrix formulation of SVM is presented as follows:
\begin{equation}
\label{eqn1}
    \begin{split}
        & \underset{W, \, Q}{min} \quad \frac{1}{2}Tr(W^{T}W) \,\, + \,\, c \, Tr(E^{T}Q), \\
        & s.t. \quad \quad Y\circ(XW) \,\, + \,\, Q \geq E, \\
        & \quad \quad \quad Q \geq 0.
    \end{split}
\end{equation}

Here, $W$ $\in$ $\mathcal{R}^{(d+1)\times m}$ is the coefficient matrix, $Q$ $\in$ $\mathcal{R}^{n\times m}$ is the error matrix, $E$ is the matrix of all ones of appropriate dimensions and $\circ$ is the Hadamard product.
It can be observed that the \cref{eqn1} jointly optimizes all classes to obtain respective hyperplane normals i.e. $W$, in one single optimisation. The Lagrangian can be written as:

\begin{equation}
\label{eqn2}
    \begin{split}
        & \underset{}{L(W,  Q)} \,\, = \quad \frac{1}{2}Tr(W^{T}W) \,\, + \,\, c \, Tr(E^{T}Q)\\              & \quad \quad \quad - \,\, Tr(\alpha^{T}( Y\circ(XW) \,\, + \,\, Q \,\,- \,\,E))\\
        & \quad \quad \quad -\,\,Tr(\beta^{T}Q) .
    \end{split}
\end{equation}
Applying the principles of matrix calculus for differentiation \cite{petersen2008matrix}, we get:
\begin{equation}
    \label{eqn3}
    \frac{\partial L}{\partial W} \,\,=\,\, W \,\,-\,\, X^{T}(\alpha \circ Y) \,\,=\,\, 0 \,\, \implies W\,\,=\,\,X^{T}(\alpha \circ Y),
\end{equation}
\begin{equation}
\label{eqn4}
    \frac{\partial L}{\partial Q} \,\,=\,\, cE\,\,-\,\, \alpha \,\, - \beta\,\, = 0\,\, \implies cE\,\,=\,\,\alpha \,\, +\,\, \beta .
\end{equation}
 Substituting \cref{eqn4} back into \cref{eqn2}, we get:
\begin{equation}
\label{eqn5}
    \begin{split}
        & \underset{}{L} \,\, = \quad \frac{1}{2}Tr(W^{T}W)\,\, - \,\, Tr(\alpha^{T}( Y\circ(XW))) \,\, + \,\,Tr(\alpha^{T}E).
    \end{split}
\end{equation}
Using $Tr(A^{T})=Tr(A)$, we get:
\begin{equation}
\label{eqn6}
    \begin{split}
        & \underset{}{L} \,\, = \quad \frac{1}{2}Tr(W^{T}W) \,\, - \,\, Tr((XW)^{T}(Y \circ \alpha)) \,\, + \,\,Tr(E^{T}\alpha).
    \end{split}
\end{equation}
Substituting \cref{eqn3} back into \cref{eqn6}, we get:
\begin{equation}
\label{eqn7}
    \begin{split}
        & \underset{}{L} \,\, = \,\, \frac{1}{2}Tr( (\alpha \circ Y)XX^{T}(\alpha \circ Y) )\,\, - \,\,Tr(E^{T}\alpha).
    \end{split}
\end{equation}
$XX^{T}$ is the linear kernel, which can also be replaced by any kernel function $K(\cdot , \cdot)$. Note that since $X$ is augmented by column vector of ones, $XX^{T} \,\,=\,\, K(X,X) \,\,+\,\, E \,\, = \,\, \Bar{K}(X,X)$. 

Finally, We obtain the Matrix optimisation of SVM, which is:
\begin{equation}
\label{eqn8}
    \begin{split}
        & \underset{\alpha}{min} \,\,L\,\, = \,\, \frac{1}{2}Tr( (\alpha \circ Y)^{T}\Bar{K}(\alpha \circ Y) )\,\, - \,\,Tr(E^{T}\alpha),\\
        & s.t. \quad \quad \quad 0\,\, \leq\,\, \alpha \,\, \leq \,\, cE.
    \end{split}
\end{equation}

\textbf{Calculation of Gradient}
\begin{equation}
\label{eqn9}
    \nabla_{\alpha} L\,\,=\,\, ( \,\,\Bar{K}(\alpha \circ Y)\,\,)\circ Y \,\,-\,\,E.
\end{equation}
\textbf{Calculating Lipchitz constant}\\
  $\norm{ \nabla L_{1} - \nabla L_{2} }$   $\leq$   $L_{f}\norm{\Delta \alpha}$,\\
  $\norm{ \nabla L_{1} - \nabla L_{2} } \,\, \leq\,\, \norm{(\,\Bar{K}(\alpha_{1} \circ Y)\,)\circ Y\,\,-\,\,(\,\Bar{K}(\alpha_{2} \circ Y)\,)\circ Y}^{2}_{F}$,\\
  $\implies \norm{\Bar{K}( \Delta\alpha \circ Y)\circ Y}^{2}_{F}$
  $\implies \norm{\Bar{K}}^{2}_{F} \norm{( \Delta\alpha \circ Y)}^{2}_{F}  \norm{Y}^{2}_{F}$ 
  $\implies \norm{\Bar{K}}^{2}_{F} \norm{\Delta\alpha}^{2}_{F} \norm{Y}^{2}_{F}  \norm{Y}^{2}_{F}$ 
  $\implies \norm{\Bar{K}}^{2}_{F} \norm{\Delta\alpha}^{2}_{F}$, \\
  $L_{f}\,\,=\,\, \sqrt{\norm{\Bar{K}}^{2}_{F}}\,\,=\,\,\norm{\Bar{K}}_{
  F}$.

\subsection{Algorithm}
\label{Algorithm}
The optimisation \cref{eqn8} can be efficiently solved by gradient-based methods. We choose to solve the same by employing the AGD method. It should be noted that AGD has convergence rate $\mathcal{O}(\frac{1}{k^{2}})$ for the first-order method \cite{nesterov1983method}. The algorithm for the Matrix-SVM is presented in \cref{algo1}
\begin{algorithm}
\caption{Accelerated Gradient Descend For solving Matrix-SVM (\cref{eqn8}) }
\label{algo1}
\begin{algorithmic}
    \State Input Training Information: $X\,\, \in \mathcal{R}^{n\times d}$, $Y\,\, \in \,\, \{1,-1\}^{n \times m}$, trade-off hyperparameter c
    \State Input Testing Information: $X_{n_{t}}\,\, \in \mathcal{R}^{n_{t}\times d}$
    \State Output: $\alpha$, $W$, Predicted label Matrix
    \State Initialise: Learning rate $\mu \,=\, \frac{1}{L_{f}}$
    \State Initialise: $\alpha_{1},\delta$ as Zero matrix
    \State Initialise: $t \,= \,1$, $z_1 \, = \, 1$
    \While{$\alpha$ not converged}
    
    \State $\alpha_{t+1}$ $\leftarrow$ $\delta$ - $\mu\,\nabla_{\delta} L(\delta)$ \Comment Calculate the Gradient of objective function $\nabla L(\delta)$  using \cref{eqn9}
    \State Clip $\alpha_{t+1}$ as: $0 \,\,\ \leq \,\, \alpha_{t+1} \,\, \leq \,\, c\,E$
    \State $z_{t+1}$ $\leftarrow$ $\frac{1\,+\, \sqrt{1\,+\,4z_{t}^{2}}}{2}$
    \State $\delta$ $\leftarrow$ $\alpha_{t+1}$ + $\frac{z_{t}\,-\,1}{z_{t+1}}(\alpha_{t+1} \, - \, \alpha_{t})$ 
    \State Clip $\delta$ as: $0 \,\,\ \leq \,\, \delta \,\, \leq \,\, c\,E$
    \State $t\, \leftarrow \,t+1$
    \EndWhile
    \State Calculate $W$ using \cref{eqn3}
    \State Predicted label Matrix: $X_{n_{t}}W$ = $(\,\Bar{K}(X_{n_{t}}, \, X)\,) (\alpha \circ Y)$
    \end{algorithmic}
\end{algorithm}

\subsection{Connection with existing approaches}
\label{Connection with existing approaches}
In this section, we delve into the relationship between Matrix-SVM and several existing models that can be challenging to conceptualise using vector notations. By employing Matrix-SVM, we are able to establish a connection and provide a clearer understanding of these models, thus enhancing their interpretability and facilitating meaningful insights.

\subsubsection{Connection with Multi-label Relationship Learning}
\label{Connection with Multi-label Relationship Learning}
The optimization approach described in \cite{zhang2013multilabel} is similar to the one presented here (i.e. \cref{eqn10a}), as it involves solving the same dual problem. However, the authors utilize vector optimization to formulate their approach. Additionally, this method can be considered an extension of Matrix-SVM since it introduces modifications to the output ($XW$) by effectively incorporating the label correlation $A$. 
\begin{equation}
\label{eqn10a}
    \begin{split}
        & \underset{W, \, Q}{min} \quad \frac{1}{2}Tr(W^{T}W) \,\, + \,\, c \, Tr(E^{T}Q) ,\\
        & s.t. \quad \quad Y\circ(XWA) \,\, + \,\, Q \geq E ,\\
        & \quad \quad \quad Q \geq 0.
    \end{split}
\end{equation}

\subsubsection{Connection with Similarity-Based Classification Approaches}
\label{Connection with Similarity-Based Classification Approaches}
The solution (Weight matrix $W$) to \cref{eqn10a} can be written as:
\begin{equation}
\label{eqn13}
\begin{split}
    & W = X^{T}(\alpha \circ Y)A, \\
    & XW = XX^{T}(\alpha \circ Y)A ,\\
    & Z = \bar{K}(X,X)*(\alpha \circ Y)A.
\end{split}
\end{equation}
When considering the case where $\alpha = E$, the given expression simplifies as ($Z \,\,-\,\, K(X, X)*(Y)A$), which requires sample and label similarity matrices only. This expression has found wide application in enhancing limited supervision challenges within multilabel learning, including tasks such as Semi-Supervised Partial Multi-label Learning \cite{xie2020semi} and Semi-Supervised Weak Label Learning \cite{dong2018learning}. 

Therefore, it can be concluded that similarity-based learning frameworks can be viewed as extensions of SVM, as they build upon the foundational concepts of SVM and incorporate this term to tackle various challenges in multilabel learning.

\subsubsection{Connection with Limited Supervision Challenges}
\label{Connection with Limited Supervision Challenges}
The inherent nature of SVM leads to sparse dual solution $\alpha$. Consequently, the term ($\alpha \circ Y$) indicates that a substantial portion of the labeled information in $Y$ is not required. Interestingly, this observation reveals that only specific labeled information plays a crucial role in learning $W$.
This insight opens up avenues for addressing other challenges related to limited supervision. For example, in multilabel learning with missing labels, only the essential label information can be obtained instead of recovering the full label matrix $Y$. To add more, enforcing row sparsity in $\alpha$ can prove beneficial in addressing Semi-Supervised Weak Label learning.

\subsubsection{Connection with Least Squares loss SVM}
\label{Connection with Least Squares loss SVM}
The primal of Matrix SVM \cref{eqn1}, using the quadratic loss, can be reformulated as:
\begin{equation}
    \label{eqn10}
    \begin{split}
        & \underset{W, \, Q}{min} \quad \frac{1}{2}Tr(W^{T}W) \,\, + \,\, c \, Tr(Q^{T}Q), \\
        & s.t. \quad \quad Y\circ(XW) \,\, + \,\, Q \,\,=\,\, E. \\
    \end{split}
\end{equation}
Further simplifying the constraint for $Q$:
\begin{equation}
    \label{eqn11}
    \begin{split}
        &   Q \,\,=\,\, E \,\,-\,\, Y\circ(XW),\\
        &   Q \,\,=\,\, Y\circ Y \,\,-\,\, Y\circ(XW),\\
        &   Q \,\,=\,\, Y\circ (Y \,\,-\,\, XW),\\
        &   \norm{Q}^{2}_{F} \,\,=\,\,\norm{Y\circ (Y \,\,-\,\, XW)}^{2}_{F},\\   
        &   \norm{Q}^{2}_{F} \,\,=\,\, \norm{Y}^{2}_{F} \norm{Y \,\,-\,\, XW}^{2}_{F}, \\
        &   \norm{Q}^{2}_{F} \,\,\simeq\,\, \norm{Y \,\,-\,\, XW}^{2}_{F}.
    \end{split}
\end{equation}
Plugging back into \cref{eqn10}, we get:
\begin{equation}
    \label{eqn12}
    \begin{split}
        & \underset{W, \, Q}{min} \quad \frac{1}{2}Tr(W^{T}W) \,\, + \,\, c \, \norm{Y \,\,-\,\, XW}^{2}_{F}.
    \end{split}
\end{equation}
The equation presented in \cref{eqn12} serves as a fundamental framework widely adapted to tackle various challenges in multilabel learning. Consequently, it can be asserted that a significant number of multilabel models can be regarded as extensions of the Matrix-SVM, albeit with modifications to suit specific requirements.

\section{Experiments}
\label{Experiments}
The experiments are performed on well-known diverse datasets (i.e. in accordance with size as well as domain, reported in \cref{multi_class_descrp} and \cref{multi_label_descrp}) using ten-fold cross-validation in MATLAB \cite{matlab2012matlab} version 9.4 under Microsoft Windows environment on a machine with 3.40 GHz i7 CPU and 16 GB RAM. The value of user-defined parameter $c$, is fixed at $c = 1$. 
All datasets are normalised in the range $[1,-1]$. All comparing models are implemented in Matlab. 
For measuring the time of the algorithms, we use tic-toc (Matlab). The training time is reported in seconds.
 We use RBF kernel ($K(x_1,x_2) = e^{-p\norm{x_1 - x_2}^{2}}$) for all kernelised models, $p$ fixed at $0.7$ for multiclass datasets and $0.3$ for multilabel datasets. The characteristics of the datasets are reported as samples $\times$ features $\times$ classes/labels in \cref{multi_class_main} and \cref{multi_label_main}. Since Satimage, Shuttle and Mediamill are very large datasets, hence a random of 4000 samples are selected.

\begin{table*}[]
    \centering
     \caption{Description on Multi-class datasets}
    \label{multi_class_descrp}

\begin{tabular}{|l|l|l|l|} \hline
\textbf{Dataset} & \textbf{Size} & \textbf{samples $\times$ features $\times$ classes} & \textbf{Domain}\\ \hline
Ecoli            & Small & 338 $\times$ 7 $\times$ 8 & Biology      \\ \hline
Yeast-multiclass & Medium & 484 $\times$ 8 $\times$ 10 & Biology    \\ \hline
Segment          & Medium & 2310 $\times$ 19 $\times$ 7 & Image    \\ \hline
Satimage         & Large  & 4000 $\times$ 36 $\times$ 6 & Image     \\ \hline
Shuttle          & Large  & 4000 $\times$ 9 $\times$ 7 & Other    \\ \hline
\end{tabular}

\end{table*}

\begin{table}
    \centering
     \caption{Description on Multi-label datasets}
    \label{multi_label_descrp}

\begin{tabular}{|l|l|l|l|} \hline
\textbf{Dataset} & \textbf{Size} & \textbf{samples $\times$ features $\times$ classes} & \textbf{Domain}\\ \hline
Emotions         & Small & 593 $\times$ 72 $\times$ 6  & Music    \\ \hline
Genbase          & Small  & 662 $\times$ 1185 $\times$ 27 & Biology   \\ \hline
Languagelog      & Medium  & 1460 $\times$ 1004 $\times$ 75 & Text  \\ \hline
Yeast            & Medium   & 2417 $\times$ 103 $\times$ 14 & Biology \\ \hline
Mediamill        & Large    & 4000 $\times$ 120 $\times$ 101 & Video \\ \hline

\end{tabular}
\end{table}

\begin{table*}
     \caption{Results on Multi-class datasets}
    \label{multi_class_main}
    \centering
    \resizebox{\linewidth}{!}{
    \begin{tiny}

\begin{tabular}{|l|c|c|c|c|}
\hline

&\cellcolor{gray!35}\textbf{BR-SVM(L)}&\cellcolor{gray!35}\textbf{M-SVM(L)}&\cellcolor{gray!35}\textbf{BR-SVM(RBF)}&\cellcolor{gray!35}\textbf{M-SVM(RBF)}\\\hline

\multicolumn{5}{|c|}{\cellcolor{gray!35}\textbf{Ecoli ( 338 $\times$ 7 $\times$ 8) }}\\ \hline 
\cellcolor{gray!35}\textbf{Time ($\downarrow$)}&0.094±0.008&\cellcolor{gray!35}\textbf{0.028±0.002}&0.069±0.006&\cellcolor{gray!35}\textbf{0.023±0.002}\\\hline
\cellcolor{gray!35}\textbf{ExactMatch ($\uparrow$)}&0.706±0.016&0.718±0.021&0.833±0.016&0.83±0.018\\\hline
\cellcolor{gray!35}\textbf{HammingLoss ($\downarrow$)}&0.05±0.004&0.046±0.004&0.033±0.003&0.033±0.004\\\hline
\cellcolor{gray!35}\textbf{MacroF1 ($\uparrow$)}&0.474±0.02&0.504±0.023&0.544±0.032&0.542±0.034\\\hline
\cellcolor{gray!35}\textbf{MicroF1 ($\uparrow$)}&0.793±0.015&0.806±0.016&0.865±0.012&0.863±0.015\\\hline
\cellcolor{gray!35}\textbf{AvgPrecision ($\uparrow$)}&0.832±0.011&0.837±0.014&0.886±0.015&0.878±0.016\\\hline

\multicolumn{5}{|c|}{\cellcolor{gray!35}\textbf{Yeast-multiclass ( 1484 $\times$ 8 $\times$ 10) }}\\ \hline 
\cellcolor{gray!35}\textbf{Time ($\downarrow$)}&1.021±0.027&\cellcolor{gray!35}\textbf{0.566±0.026}&0.767±0.012&\cellcolor{gray!35}\textbf{0.428±0.013}\\\hline
\cellcolor{gray!35}\textbf{ExactMatch ($\uparrow$)}&0.189±0.009&0.189±0.009&0.299±0.017&0.301±0.017\\\hline
\cellcolor{gray!35}\textbf{HammingLoss ($\downarrow$)}&0.087±0.001&0.087±0.001&0.082±0.001&0.082±0.001\\\hline
\cellcolor{gray!35}\textbf{MacroF1 ($\uparrow$)}&0.265±0.025&0.262±0.024&0.322±0.029&0.324±0.03\\\hline
\cellcolor{gray!35}\textbf{MicroF1 ($\uparrow$)}&0.304±0.012&0.303±0.013&0.419±0.016&0.421±0.016\\\hline
\cellcolor{gray!35}\textbf{AvgPrecision ($\uparrow$)}&0.302±0.007&0.301±0.008&0.395±0.014&0.396±0.014\\\hline

\multicolumn{5}{|c|}{\cellcolor{gray!35}\textbf{Segment ( 2310 $\times$ 19 $\times$ 7) }}\\ \hline 
\cellcolor{gray!35}\textbf{Time ($\downarrow$)}&9.912±0.074&\cellcolor{gray!35}\textbf{5.69±0.035}&1.928±0.025&\cellcolor{gray!35}\textbf{0.967±0.015}\\\hline
\cellcolor{gray!35}\textbf{ExactMatch ($\uparrow$)}&0.83±0.005&0.828±0.003&0.913±0.004&0.914±0.004\\\hline
\cellcolor{gray!35}\textbf{HammingLoss ($\downarrow$)}&0.03±0.001&0.029±0.001&0.017±0.001&0.016±0.001\\\hline
\cellcolor{gray!35}\textbf{MacroF1 ($\uparrow$)}&0.878±0.005&0.875±0.006&0.935±0.004&0.938±0.005\\\hline
\cellcolor{gray!35}\textbf{MicroF1 ($\uparrow$)}&0.891±0.003&0.892±0.003&0.94±0.003&0.942±0.004\\\hline
\cellcolor{gray!35}\textbf{AvgPrecision ($\uparrow$)}&0.935±0.003&0.936±0.003&0.966±0.002&0.968±0.003\\\hline

\multicolumn{5}{|c|}{\cellcolor{gray!35}\textbf{Satimage ( 4000 $\times$ 36 $\times$ 6) }}\\ \hline 
\cellcolor{gray!35}\textbf{Time ($\downarrow$)}&13.193±0.104&\cellcolor{gray!35}\textbf{4.882±0.044}&7.87±0.054&\cellcolor{gray!35}\textbf{2.661±0.013}\\\hline
\cellcolor{gray!35}\textbf{ExactMatch ($\uparrow$)}&0.737±0.007&0.738±0.006&0.882±0.003&0.88±0.003\\\hline
\cellcolor{gray!35}\textbf{HammingLoss ($\downarrow$)}&0.054±0.001&0.054±0.001&0.03±0.001&0.031±0.001\\\hline
\cellcolor{gray!35}\textbf{MacroF1 ($\uparrow$)}&0.687±0.005&0.688±0.004&0.875±0.003&0.871±0.003\\\hline
\cellcolor{gray!35}\textbf{MicroF1 ($\uparrow$)}&0.822±0.005&0.823±0.004&0.907±0.003&0.906±0.003\\\hline
\cellcolor{gray!35}\textbf{AvgPrecision ($\uparrow$)}&0.867±0.004&0.868±0.003&0.934±0.002&0.932±0.002\\\hline

\multicolumn{5}{|c|}{\cellcolor{gray!35}\textbf{Shuttle ( 4000 $\times$ 9 $\times$ 7) }}\\ \hline 
\cellcolor{gray!35}\textbf{Time ($\downarrow$)}&6.751±0.05&\cellcolor{gray!35}\textbf{3.546±0.018}&6.172±0.05&\cellcolor{gray!35}\textbf{2.817±0.031}\\\hline
\cellcolor{gray!35}\textbf{ExactMatch ($\uparrow$)}&0.829±0.005&0.837±0.004&0.997±0.001&0.997±0.001\\\hline
\cellcolor{gray!35}\textbf{HammingLoss ($\downarrow$)}&0.027±0.001&0.025±0.001&0.001±0&0.001±0\\\hline
\cellcolor{gray!35}\textbf{MacroF1 ($\uparrow$)}&0.283±0&0.283±0&0.427±0.001&0.427±0.001\\\hline
\cellcolor{gray!35}\textbf{MicroF1 ($\uparrow$)}&0.899±0.003&0.904±0.003&0.997±0.001&0.997±0.001\\\hline
\cellcolor{gray!35}\textbf{AvgPrecision ($\uparrow$)}&0.876±0.003&0.875±0.003&0.998±0.001&0.997±0.001\\\hline
\end{tabular}
\end{tiny}
}
\end{table*}

\begin{table*}
    \centering
     \caption{Results on Multi-label datasets}
    \label{multi_label_main}
    \resizebox{\linewidth}{!}{
    \begin{tiny}
\begin{tabular}{|l|c|c|c|c|}
\hline
&\cellcolor{gray!35}\textbf{BR-SVM(L)}&\cellcolor{gray!35}\textbf{M-SVM(L)}&\cellcolor{gray!35}\textbf{BR-SVM(RBF)}&\cellcolor{gray!35}\textbf{M-SVM(RBF)}\\\hline

\multicolumn{5}{|c|}{\cellcolor{gray!35}\textbf{Emotions ( 593 $\times$ 72 $\times$ 6) }}\\ \hline 
\cellcolor{gray!35}\textbf{Time ($\downarrow$)}&0.273±0.013&\cellcolor{gray!35}\textbf{0.128±0.01}&0.058±0.008&\cellcolor{gray!35}\textbf{0.035±0.004}\\\hline
\cellcolor{gray!35}\textbf{ExactMatch ($\uparrow$)}&0.262±0.027&0.27±0.028&0.3±0.012&0.307±0.012\\\hline
\cellcolor{gray!35}\textbf{HammingLoss ($\downarrow$)}&0.208±0.008&0.206±0.008&0.183±0.005&0.181±0.005\\\hline
\cellcolor{gray!35}\textbf{MacroF1 ($\uparrow$)}&0.619±0.016&0.616±0.016&0.634±0.014&0.639±0.014\\\hline
\cellcolor{gray!35}\textbf{MicroF1 ($\uparrow$)}&0.643±0.017&0.642±0.018&0.668±0.013&0.673±0.013\\\hline
\cellcolor{gray!35}\textbf{AvgPrecision ($\uparrow$)}&0.742±0.01&0.739±0.011&0.762±0.01&0.765±0.01\\\hline

\multicolumn{5}{|c|}{\cellcolor{gray!35}\textbf{Genbase ( 662 $\times$ 1185 $\times$ 27) }}\\ \hline 
\cellcolor{gray!35}\textbf{Time ($\downarrow$)}&0.292±0.013&\cellcolor{gray!35}\textbf{0.247±0.025}&0.268±0.009&\cellcolor{gray!35}\textbf{0.062±0.006}\\\hline
\cellcolor{gray!35}\textbf{ExactMatch ($\uparrow$)}&0.952±0.01&0.983±0.005&0.829±0.014&0.828±0.015\\\hline
\cellcolor{gray!35}\textbf{HammingLoss ($\downarrow$)}&0.002±0&0.001±0&0.01±0.001&0.01±0.001\\\hline
\cellcolor{gray!35}\textbf{MacroF1 ($\uparrow$)}&0.598±0.02&0.661±0.016&0.522±0.03&0.521±0.03\\\hline
\cellcolor{gray!35}\textbf{MicroF1 ($\uparrow$)}&0.978±0.005&0.993±0.002&0.883±0.014&0.883±0.014\\\hline
\cellcolor{gray!35}\textbf{AvgPrecision ($\uparrow$)}&0.978±0.006&0.995±0.001&0.847±0.014&0.847±0.014\\\hline

\multicolumn{5}{|c|}{\cellcolor{gray!35}\textbf{Languagelog ( 1460 $\times$ 1004 $\times$ 75) }}\\ \hline 
\cellcolor{gray!35}\textbf{Time ($\downarrow$)}&8.022±0.034&\cellcolor{gray!35}\textbf{4.707±0.054}&2.728±0.044&\cellcolor{gray!35}\textbf{0.485±0.018}\\\hline
\cellcolor{gray!35}\textbf{ExactMatch ($\uparrow$)}&0.157±0.008&0.155±0.008&0.141±0.007&0.141±0.007\\\hline
\cellcolor{gray!35}\textbf{HammingLoss ($\downarrow$)}&0.184±0.003&0.211±0.004&0.184±0.004&0.184±0.004\\\hline
\cellcolor{gray!35}\textbf{MacroF1 ($\uparrow$)}&0.353±0.008&0.385±0.007&0.104±0.003&0.111±0.002\\\hline
\cellcolor{gray!35}\textbf{MicroF1 ($\uparrow$)}&0.532±0.005&0.511±0.005&0.398±0.008&0.416±0.007\\\hline
\cellcolor{gray!35}\textbf{AvgPrecision ($\uparrow$)}&0.516±0.006&0.494±0.006&0.533±0.007&0.543±0.006\\\hline

\multicolumn{5}{|c|}{\cellcolor{gray!35}\textbf{Yeast ( 2417 $\times$ 103 $\times$ 14 ) }}\\ \hline 
\cellcolor{gray!35}\textbf{Time ($\downarrow$)}&8.353±0.075&\cellcolor{gray!35}\textbf{3.168±0.062}&3.451±0.039&\cellcolor{gray!35}\textbf{1.222±0.043}\\\hline
\cellcolor{gray!35}\textbf{ExactMatch ($\uparrow$)}&0.147±0.006&0.149±0.009&0.2±0.012&0.204±0.011\\\hline
\cellcolor{gray!35}\textbf{HammingLoss ($\downarrow$)}&0.2±0.002&0.2±0.002&0.185±0.003&0.185±0.003\\\hline
\cellcolor{gray!35}\textbf{MacroF1 ($\uparrow$)}&0.327±0.003&0.327±0.003&0.384±0.006&0.385±0.006\\\hline
\cellcolor{gray!35}\textbf{MicroF1 ($\uparrow$)}&0.635±0.004&0.635±0.004&0.657±0.005&0.657±0.006\\\hline
\cellcolor{gray!35}\textbf{AvgPrecision ($\uparrow$)}&0.672±0.004&0.671±0.004&0.683±0.005&0.684±0.005\\\hline

\multicolumn{5}{|c|}{\cellcolor{gray!35}\textbf{Mediamill ( 4000 $\times$ 120 $\times$ 101) }}\\ \hline 
\cellcolor{gray!35}\textbf{Time ($\downarrow$)}&253.649±1.814&\cellcolor{gray!35}\textbf{59.931±0.385}&90.031±0.815&\cellcolor{gray!35}\textbf{21.331±0.144}\\\hline
\cellcolor{gray!35}\textbf{ExactMatch ($\uparrow$)}&0.082±0.005&0.083±0.005&0.093±0.005&0.094±0.005\\\hline
\cellcolor{gray!35}\textbf{HammingLoss ($\downarrow$)}&0.03±0&0.03±0&0.03±0&0.03±0\\\hline
\cellcolor{gray!35}\textbf{MacroF1 ($\uparrow$)}&0.037±0&0.037±0.001&0.044±0.001&0.044±0.001\\\hline
\cellcolor{gray!35}\textbf{MicroF1 ($\uparrow$)}&0.536±0.004&0.535±0.004&0.546±0.004&0.546±0.004\\\hline
\cellcolor{gray!35}\textbf{AvgPrecision ($\uparrow$)}&0.514±0.004&0.514±0.004&0.519±0.003&0.519±0.003\\\hline

\end{tabular}
\end{tiny}
}
\end{table*}

 \subsection{Discussion on Results}
 \label{Discussion on Results}
 In \cref{multi_class_main} and \cref{multi_label_main}, the first and second column report the results for the linear kernel, while the remaining column report results for the RBF kernel.
Analysis of \cref{multi_class_main} and \cref{multi_label_main} reveals that Matrix SVM (M-SVM) achieves results similar to BR-SVM. However, a significant disparity in training time is observed even though the same method (AGD) is employed for solving both models. This discrepancy can be attributed to the fact that Matrix SVM addresses the unified problem, while BR-SVM independently solves the optimization for each label. Moreover, the joint optimization approach of Matrix SVM enhances its stability compared to BR-SVM. 

In the case of medium and large datasets for both multiclass and multilabel datasets, it can be clearly observed that kernelised versions ( i.e. BR-SVM(RBF) and M-SVM(RBF) ) are much faster than their respective counterparts ( i.e. BR-SVM(L) and M-SVM(L) ). This is due to the fact that kernelised models are able to achieve faster convergence than linear models. Generally, the kernel trick projects the data into higher dimensions which results in better accuracy than the linear models. But in the case of high-dimensionality datasets ( i.e. sample size $\simeq$ feature size, here Genbase and Languagelog ), kernelisation is not necessarily required. Hence, the linear models have greater accuracy than kernel models.

\section{Conclusions}
\label{Conclusions}
We strongly advocate for the adoption of our proposed Matrix formulation for Support Vector Machines as a new baseline for comparing against emerging multilabel, multiclass algorithms. Through extensive experiments, we demonstrate that Matrix SVM achieves comparable results to BR-SVM while exhibiting superior time efficacy. By optimizing all classes within a single optimization process, our matrix formulation ensures a more stable and reliable optimization. Moreover, the matrix formulation uncovers essential insights and advantages that might go unnoticed in traditional vector-based notations. We emphasize that many multilabel models can be viewed as extensions of SVM, with tailored modifications to cater to specific requirements. The matrix formulation presented in this paper provides a good foundation, for developing more sophisticated models capable of effectively addressing the unique challenges encountered in multilabel learning. 
In our upcoming research, we intend to extend the Matrix-SVM to tackle the challenge of multilabel learning with missing labels while simultaneously exploiting label correlations. By doing so, we aim to develop an enhanced framework that can effectively handle missing labels and leverage label correlations for improved multilabel learning performance.

\textbf{Acknowledgements}\\
 We would like to express gratitude to Dr Pritam Anand for reviewing the manuscript in the initial stage and for providing constructive feedback which helped to improve the manuscript.

\medskip

\end{document}